\documentclass[journal]{IEEEtran}

\usepackage{cite}

\usepackage{chngcntr}
\usepackage[utf8]{inputenc} % allow utf-8 input
\usepackage[T1]{fontenc}    % use 8-bit T1 fonts
\usepackage{hyperref}       % hyperlinks
\usepackage{url}            % simple URL typesetting
\usepackage{booktabs}       % professional-quality tables
\usepackage{amsfonts}       % blackboard math symbols
\usepackage{nicefrac}       % compact symbols for 1/2, etc.
\usepackage{microtype}      % microtypography
\usepackage{xcolor}         % colors
\usepackage{fixltx2e}
\usepackage{amsmath}
\usepackage{amsthm}
\usepackage{ulem}
\usepackage{graphicx}
\usepackage{subcaption}
\usepackage{enumitem}
\usepackage{makecell}

\newtheorem{theorem}{Theorem}
\newtheorem{lemma}[theorem]{Lemma}

\newcommand{\bfp}{\mathbf{p}}
\newcommand{\bft}{\mathbf{t}}

\newcommand{\bff}{\mathbf{f}}

\newcommand{\bfy}{\mathbf{y}}

\newcommand{\bfK}{\mathbf{K}}
\newcommand{\bfI}{\mathbf{I}}

\newcommand{\bfA}{\mathbf{A}}
\newcommand{\bfD}{\mathbf{D}}
\newcommand{\bfW}{\mathbf{W}}
\newcommand{\bfS}{\mathbf{S}}
\newcommand{\bfT}{\mathbf{T}}
\newcommand{\bfL}{\mathbf{L}}
\newcommand{\bfP}{\mathbf{P}}

\newcommand{\CC}{\mathcal{C}}

\usepackage{tikz}

\usetikzlibrary{arrows,shapes}

\definecolor{mygreen}{rgb}{0,0.6,0}

\pgfdeclarelayer{background}
\pgfsetlayers{background,main}

\tikzstyle{vertex}=[circle,fill=black!25,minimum size=20pt,inner sep=0pt]
\tikzstyle{selected vertex} = [vertex, fill=red!24]
\tikzstyle{select vertex} = [vertex, fill=blue!24]
\tikzstyle{selectx vertex} = [vertex, fill=green!24]
\tikzstyle{edge} = [draw,thick,-]
\tikzstyle{selected edge} = [draw,line width=5pt,-,red!50]

\usepackage{relsize}
\usepackage{bm}

\usetikzlibrary{positioning,decorations.pathmorphing}

\definecolor{mygreen}{rgb}{0,0.6,0}
\definecolor{mymauve}{rgb}{0.58,0,0.82} 
\definecolor{camdrk}{RGB}{0,62,114}

\begin{document}

\title{Haar Wavelet Feature Compression for Quantized Graph Convolutional Networks}
\author{Moshe Eliasof, Benjamin Bodner, and Eran Treister
\thanks{The authors are with the Department
of Computer Science at Ben-Gurion University of the Negev, Beer-Sheva, Israel. e-mails: (eliasof@post.bgu.ac.il, benjybo7@gmail.com, erant@cs.bgu.ac.il).}% <-this % stops a space
}

\markboth{}%
{Shell \MakeLowercase{\textit{et al.}}: Bare Demo of IEEEtran.cls for IEEE Journals}

\maketitle
\begin{abstract}
Graph Convolutional Networks (GCNs) are widely used in a variety of applications, and can be seen as an unstructured version of standard Convolutional Neural Networks (CNNs). As in CNNs, the computational cost of GCNs for large input graphs (such as large point clouds or meshes) can be high and inhibit the use  of these networks, especially in environments with low computational resources. To ease these costs, quantization can be applied to GCNs. However, aggressive quantization of the feature maps can lead to a significant degradation in performance. On a different note, Haar wavelet transforms are known to be one of the most effective and efficient approaches to compress signals. Therefore, instead of applying aggressive quantization to feature maps, we propose to utilize Haar wavelet compression and light quantization to reduce the computations and the bandwidth involved with the network. We demonstrate that this approach surpasses aggressive feature quantization by a significant margin, for a variety of problems ranging from node classification to point cloud classification and part and semantic segmentation.
\end{abstract}

\begin{IEEEkeywords}
Graph Convolutional Networks, Graph Wavelet Transform, Quantized Neural Networks, Network compression.
\end{IEEEkeywords}

\IEEEpeerreviewmaketitle

\section{Introduction}
\IEEEPARstart{G}{raph} convolutional networks (GCNs) have been shown to be highly successful when applied to a wide array of problems and domains, including social analysis \cite{qiu2018deepinf, li2019encoding}, recommendation systems \cite{ying2018graph}, computational biology \cite{eliasof2021mimetic} and computer vision and graphics \cite{monti2017geometric, wang2018dynamic, eliasof2020diffgcn}. Conceptually, GCNs are an unstructured version of standard convolutional neural networks (CNNs) where instead of a 2D or a 3D grid, we have an unstructured graph or a mesh \cite{eliasof2020diffgcn}.

The computational cost of GCNs, as with structured CNNs, is directly affected by the size of their inputs and the intermediate feature-maps (activations) throughout the network. In many real-world applications, such as LiDAR-based point cloud segmentation, large point clouds or graphs are required during training and deployment, leading to high computational and memory costs. These tremendous costs inhibit the use and deployment of these networks, especially in environments with low computational resources such as smartphones, autonomous vehicles, and specialized edge devices.

A variety of approaches have been proposed to reduce the computational costs of neural networks in recent years. Among the popular approaches are weight pruning or sparsification \cite{luo2017thinet,wen2016learning}. Such methods reduce the number of non-zero entries in the weight tensors, intending to perform computations on the non-zero entries only. However, to benefit from such pruning methods, the sparsification needs to have some structure \cite{zhou2021learning}. 

Another popular approach that we focus on in this work are quantization methods, where the numerical precision of both the weights and activations throughout the network is reduced \cite{hubara2017quantized}. This enables the operations to be carried out using low-cost fixed-point integer arithmetic, and significantly relieves the memory overhead. Recent works have demonstrated that standard CNNs can be quantized to 4-bit precision, with minimal losses in performance \cite{LSQ, li2019additive}. In addition, quantization methods were also applied to GCNs \cite{SGQuant, tailor2021degree, LPGNAS}, demonstrating similar gains in computational efficiency with minimal loss in performance.

However, in many scenarios it is evident that aggressive quantization (less than 4-bit precision) is prone to accuracy degradation. This is particularly apparent for the activation bit-width in tasks such as semantic segmentation \cite{TPAMI2020_Q_UNETS} where we wish to classify each element of the input, as opposed to classification \cite{LSQ}, where a single label is globally assigned to the input. Indeed, the majority of quantization techniques are applied and tested on image or graph classification \cite{compression_survey}. In the recent \cite{TPAMI2020_Q_UNETS}, which targets quantized U-Nets, the activations bit rates are relatively high (8 bits), while those can be low for the learnt weights (2 bits). In this work, we wish to improve activations compression in GCNs beyond quantization and to ease the bandwidth and computations of GCNs, especially for semantic segmentation on dense point clouds.

To this end, we harness wavelet transformations \cite{daubechies1992ten} which are well-known for their ability to compress images through sparse representation in the wavelet domain, as in the popular JPEG2000 format \cite{rabbani2002jpeg2000}. Following their success in images, wavelet transformations have been adapted for graph signals \cite{HAMMOND2011129}, and were later used in various works such as \cite{GWNN, DeepGWC, M-GWNN} to define spectral wavelet-based graph convolutions.   
Similarly, a spatial graph Haar wavelet transformation was suggested in \cite{ram2011generalized} based on a chain representation of the graph. Later, a similar principle was adapted and applied to GCN convolution and pooling \cite{fasthaarGCN, wang2020haar,zheng2020mathnet}. The advantage of the wavelet transforms lies in the sparsity of the transformed signals, which can be exploited for denoising and compression. Furthermore, wavelet convolutions typically have a wider receptive field than standard spatial convolutions, yet with similar computational costs \cite{GWNN}, making them favorable in terms of computational efficiency.

In this work, we wish to utilize both the Haar wavelet compression and quantization methods to obtain highly compressed GCNs, which exhibit minimal losses in performance. The Haar compression is applied to the activations, while quantization is applied to both weights and the Haar-transformed activations. This combination saves significant memory bandwidth and computations in the inference and training of the network. Our Haar transform is a parallelized version of \cite{ram2011generalized}, that is efficiently applied on GPUs, using graph clustering, and is similar to the transform in \cite{fasthaarGCN}. The transform is applied with linear complexity for each channel separately, using binary operations only (additions and subtractions solely), due to the simplicity of the Haar basis. The key component of our approach is to keep only the top $k$ elements of the transformed activations -- \emph{across all channels}, which we refer to as the wavelet shrinkage scheme, yielding a compressed tensor of the wavelet transform. The final component is the convolution that is applied to the compressed tensor. We show that the transform and shrinkage operations commute with the $1\times1$ convolution, meaning that the convolution step, which is at the core of CNNs and GCNs, can be performed on a significantly smaller input. This idea is executed along with quantization to further reduce the activation size and cost of convolution operations. 

Using the approach described above we compress the feature maps with minimal loss of information, which, as we demonstrate, provides better compression ratios and less performance degradation than having aggressive quantization on the activations. We propose two versions of our compression scheme: the first can be applied with minimal change to the network's weights (i.e., it is possible to use pre-trained weights in the spatial domain), and the other is a new neural architecture in the graph-Haar wavelet domain, that is suitable for large feature maps and requires less wavelet transforms.   
To summarize, our contribution is as follows:
\begin{itemize}
    \item We apply both quantization-aware training and hierarchical graph Haar transforms in the context of GCNs.
    \item We show that the sparsity of the Haar wavelet transform can be used for the compression of the graph channels. By applying joint channel shrinkage, the combination of the transform and shrinkage operation commute with a $1\times1$ convolution, obtaining bandwidth and computational savings using dense arithmetic only. We show that this approach surpasses aggressive quantization of the activation in both performance and efficiency.
    \item Two wavelet convolution schemes are suggested: the first can be applied to pre-trained weights of a network without Haar transforms (i.e., as a scheme for compression only), and the other applies multiple convolutions in the wavelet domain, requiring less forward and inverse transforms, but needs to be trained with the Haar transform from scratch. 
    %\item Lastly, we propose a novel convolution based on the wavelet sub-channels whose number of parameters do not depend on the graph number of nodes that can reach thousands. Because the wavelets are rather local this results in a local convolution whose field of view depends on the number of wavelet levels. 
\end{itemize}

\section{Related work}
\textbf{Graph Convolutional Networks.} 
GCNs come in two main types -- involving spatial and spectral convolutions \cite{wu2020comprehensive}. The spectral convolution is built on the graph Fourier transform, utilizing learnt low order Chebyshev polynomials over the graph Laplacian \cite{defferrard2016convolutional}. This formulation is also featured in the works \cite{kipf2016semi,chen20simple}, using first order polynomials. Such methods are mostly used for graph node classification problems. One of the most debatable issues in the literature with such methods is the over-smoothing problem that leads to degradation in performance as more layers are added. For example, the purpose of additions in \cite{chen20simple} over \cite{kipf2016semi} are to prevent over-smoothing.

Another family of GCNs targets more \emph{geometric} tasks such as classification and semantic segmentation of point clouds and meshes where the input features include locations of the points in a 3D space. See DGCNN \cite{wang2018dynamic}, MPNN \cite{gilmer2017neural}, DiffGCN \cite{eliasof2020diffgcn}, and references therein. Usually, given a point cloud, the underlying graph is produced by the k-nearest neighbor (kNN) algorithm, and the spatial convolution is defined by computing values in the graph edge space. For instance, in DGCNN neighboring features are subtracted, while in \cite{eliasof2020diffgcn,smirnov2021hodgenet} graph differential operators are used, and graph pooling is applied by graph clustering, e.g., Graclus \cite{dhillon2007weighted}. 

\textbf{Wavelet transforms, compression and convolution.}
Wavelet transforms are widely used for images, and can also be utilized to define convolutions and architectures in standard CNNs for a variety of imaging tasks--see \cite{liu2019multi,duan2017sar,huang2017wavelet}. 
Analogously, graph wavelet transforms \cite{HAMMOND2011129} has been used to define convolutions in GCNs. Similarly to the graph convolution, graph wavelet convolutions can be spectral \cite{GWNN, DeepGWC, M-GWNN}, and like the graph Fourier convolution, they primarily rely on the eigenbasis of the graph Laplacian, along with a heat kernel applied on the spectrum. Graph wavelets lead to sparse feature maps for node classification networks \cite{GWNN}. 

Of particular interest in this work is the graph Haar transform \cite{ram2011generalized} which has been designed to generate highly sparse image representations based on a 1D chain representation of the graph. Likewise, a parallel version of the Haar transform is used to define graph convolution and pooling \cite{fasthaarGCN, wang2020haar,zheng2020mathnet}, where parallel clustering methods are employed to generate a tree-based hierarchy.  
The advantage of the Haar transform lies in its simplicity and computational efficiency, as it does not require multiplications and can be carried out using efficient binary operations. We exploit the sparsity of the transformed features to save computations by compressing them, and while here we use the Haar transform, the standard wavelet transform \cite{GWNN} can also be used in our framework.    

\textbf{Quantized Neural Networks.} 
Research on quantization methods has been active in recent years, mostly in the context of CNNs. A plethora of quantization schemes tailored for different scenarios have been proposed, including post-training and quantization aware training schemes, uniform and non-uniform schemes, etc., see \cite{compression_survey,LSQ, li2019additive} and references therein. 
In the context GCNs, per-node bit allocation was suggested in \cite{tailor2021degree} and \cite{SGQuant}, while in \cite{LPGNAS} the bit allocation is performed for each layer, using neural architecture search methods.
In this work, we focus on the contribution of the wavelet shrinkage scheme. With that in mind, and to ensure hardware compatibility, we harness uniform quantization aware training, with per-layer quantization parameters. 

\section{Background: Neural Network Quantization and Training}
\label{sec:quantization_scheme}

The uniform quantization schemes we consider can typically be split into three steps: clip, scale and round. In the first step, the original values of the weights or activations are clipped to be within a range of values: i.e., $[-\alpha, \alpha \cdot r_{b-1}]$ for signed quantization (typically weights) and $[0, \alpha \cdot r_b]$ for unsigned quantization (typically activations\footnote{We assume that the value of the activations is non-negative (e.g., following ReLU). Hence the quantization scheme used for the activations is unsigned.}). Here, $\alpha$ is  referred to as the clipping parameter and $r_b=\frac{2^{b} - 1}{2^{b}}$, where $b$ is the number of bits. Next, the dynamic range is scaled to the target integer range of $[-2^{b-1}, 2^{b-1} - 1]$ for signed quantization and $[0, 2^{b} - 1]$ for unsigned quantization. Lastly, the values are rounded to the nearest integer.

Each layer in the neural network typically contains two clipping parameters, one for the weights and one for the activations. These parameters are optimized across all layers during the training procedure such that they yield the lowest possible loss. Several approaches have been proposed in recent years in order to optimize the clipping parameters, such as those based on quantization error minimization \cite{zhang2018lq}, backpropagation \cite{li2019additive} and more \cite{compression_survey}. Here, we base our uniform quantization aware training on \cite{li2019additive} and \cite{LSQ}, which enables the quantized network to be trained in an end-to-end manner.

To formally introduce the point-wise quantization operations we first define the quantization operator
\begin{equation} \label{quantoperator}
\textstyle    Q_{b}(x) = {\frac{\mbox{round}(2^b \cdot x)}{2^b}},
\end{equation}
where $x$ is a real-valued tensor in [-1, 1] or [0, 1] for signed or unsigned quantization, respectively. %$b$ is the number of bits that are used to represent $x$.

Given this operator, we use the re-parameterized clipping function \cite{li2019additive} to define the quantized weights and activations 
\begin{eqnarray} \label{quantweights}
    W_b &=& \alpha_W \textstyle Q_{b-1}(\mbox{clip}({\frac{W}{\alpha_W}}, -1, r_{b-1}))\\
 \label{quantactivations}
     X_b &=& \alpha_X \textstyle Q_{b}(\mbox{clip}({\frac{X}{\alpha_X}}, 0, r_b)) \label{quantacts}.
\end{eqnarray}
Here, $W, W_b$ are the original and quantized weight tensors, $X, X_b$ are the original and quantized feature tensors, and $\alpha_W, \alpha_X$ are their associated clipping parameters, respectively. 
The function $\mbox{clip}(x,a,b)$ denotes the clipping of the input $x$ onto the section $[a,b]$. Note that Eq. \eqref{quantoperator}-\eqref{quantacts} are used for training only. During inference, all weights and activations quantized, and all operations are performed using fixed-point arithmetic. %, while taking the scales into account.

Here, we also follow the common practice in quantization aware training schemes, where activations and weights are quantized during the forward pass, and in the backward pass, the Straight Through Estimator (STE) \cite{STE} is used to optimize the weights. That is, we use $W_b$ in the forward pass, but ignore $Q_b$ in the backward pass, and iterate on the floating point values of $W$ in the optimization process. Furthermore, given Eq. \eqref{quantweights}-\eqref{quantacts} the STE can also be used to calculate the gradients with respect to, for example, $\alpha_W$: 
%, \alpha_X$ \cite{LSQ}:
\begin{eqnarray} \label{weights_alpha_derrivative}
    \frac{\partial{W_b}}{\partial{\alpha_W}} = 
    \begin{cases}
    -1 & \text{if $W \leq -\alpha_W$} \\
    r_{b-1} & \text{if $W \geq \alpha_W \cdot r_{b-1}$} \\
    \frac{W_b}{\alpha_W} - \frac{W}{\alpha_W} & \text{if $-\alpha_W < W < \alpha_W \cdot r_{b-1}$} ;\\
    \end{cases}
\end{eqnarray}

The contributions are then summed over all elements of $W_b$, in order to calculate the gradient values used to update $\alpha_W$. $\alpha_X$ is updated similarly, only with respect to the activation maps. For more details, see \cite{LSQ,li2019additive}. To help the optimization we also use weight normalization before the quantization \cite{li2019additive}:
$\hat W = \frac{W - \mu}{\sigma  + \epsilon}$.
Here, $\mu$ and $\sigma$ are the mean and standard deviation of the weight tensor, respectively, and $\epsilon=10^{-6}$.

\section{Method}
\label{sec:WaveletMethod}
In this section, we describe our Haar wavelet convolution and feature compression method in detail. We aim to significantly reduce the memory bandwidth and computational cost associated with convolutions performed on intermediate feature maps. Our method is especially efficient for tasks with large feature maps (i.e., large number of elements or channels), where the size of the weights is relatively small compared to the size of the features (e.g., point cloud segmentation). Here we use the Haar transform for its computational efficiency and simplicity, such that spatial operations and fully-coupled operations like $1\times1$ convolutions are separable, where the latter is performed in the wavelet domain on a fraction of the original input size. This property is desired, as $1\times1$ convolutions are typically the most expensive component of neural networks.
 
To this end, prior to the fully-coupled convolution step, a spatial operation is applied, similar to the multiplication of input features with the graph Laplacian or gradient, which is separable between the channels---i.e., there's no coupling between the different feature maps. We then compute the graph Haar transform separately on each channel and get sparse feature maps. Then, we perform a joint shrinkage operation, in which we drop the entries with the smallest feature norms, resulting in a compressed representation of the input features. The locations of those non-zeros in the original graph are kept in a single index list (alternatively, a bit-map), as they are necessary for the inverse transform from the wavelet to the spatial domain. Ultimately,
to further improve the compression rates, we also apply relatively light quantization (8 bits) over the transformed signals, and quantize the convolution weights, as described in Sec. \ref{sec:quantization_scheme}.

We consider two versions of the compression. The first (V\textsubscript{1}), is where our transform is used for compression only, and the network itself (its weights) can in principle remain the same as without the compression, as we show in Sec. \ref{subsec:compressed_wavelet_equiv}. This version is important, for example, for scenarios where we want to compress networks without re-training (or without access to the data to train on)
using pre-trained weights. In this case, in each convolution we apply the forward and inverse Haar transforms in every convolution. In our second version (V\textsubscript{2}), we apply the non-linear activations on the compressed signals in the wavelet domain, which involves fewer Haar transforms throughout the network. Although the Haar transforms are computationally expensive, this reduces resources demand.

\subsection{Graph Haar wavelet transform}
\label{subsec:graphHarrWaveletTransform}
%\documentclass[crop, tikz]{standalone}

%\begin{document}
\begin{figure}
\begin{tikzpicture}[scale=1.8, auto,swap]
    \foreach \pos/\name in {{(0,1.5)/0}, {(2,0.5)/6}, {(4,1)/5},
                            {(0,0)/1}, {(3,0)/4},  {(3,-0.75)/3},
                             {(0.75, 0.66)/2}, {(1.5, 1.5)/7}}
     \node[vertex] (\name) at \pos {};
        
    \foreach \source/ \dest /\weight in {6/0/10, 5/6/8,1/0/5,1/6/9,
                                         4/6/7, 4/5/5,4/1/15,
                                         3/1/6,
                                         3/4/9,
                                         1/2/9,
                                         7/6/9,
                                         7/5/9,
                                         2/6/9}
        \path[edge] (\source) -- (\dest);
         
    %\foreach \vertex / \fr in {6/4}
        %\path node[selected vertex] at (\vertex) {${2}_b$};
    \foreach \vertex / \fr in {6/4,0/4, 5/4, 1/4, 4/5, 3/5, 2/5, 7/5}
        \path node[select vertex] at (\vertex) {${f}_{\vertex}$};
    \begin{pgfonlayer}{background}
        \foreach \source / \dest in {3/4, 1/2, 7/5, 0/6}
            \path[selected edge] (\source.center) -- (\dest.center);
    \end{pgfonlayer}

\end{tikzpicture}
\caption{An example of a perfect matching on a given graph (${\mathcal{E}}_p$ from Eq. \eqref{eq:subsetConditions} is labeled in red). \label{fig:pairGraph}}

\end{figure}
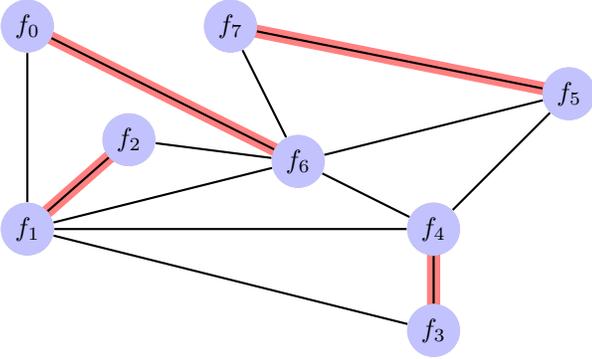

%\end{document}

%\documentclass[crop, tikz]{standalone}

%\begin{document}
\begin{figure*}
\begin{tikzpicture}[scale=1.8, auto,swap]

 %%-------------- level 2 ----------------------------------------------

    \foreach \pos/\name in {{(1,1)/0}, {(0.375,0.33)/1}, {(3,-0.375)/2},{(2.75,1.25)/3}}
     \node[vertex] (\name) at \pos {};
        
    \foreach \source/ \dest /\weight in {
                                         0/1/9,
                                         0/3/9,
                                         2/3/9,
                                         2/1/9,
                                         0/2/9,
                                         1/3/9
                                         }
        \path[edge] (\source) -- (\dest);
         
    %\foreach \vertex / \fr in {b/4}
        %\path node[selected vertex] at (\vertex) {${f}_b$};
    %\foreach \vertex / \fr in {b/4,a/4, c/4, d/4, e/5, g/5, f/5, h/5}
    %    \path node[select vertex] at (\vertex) {${f}_{\vertex}^{(1)}$};
    
    \foreach \vertex / \fr in {0/3,1/3, 2/3, 3/3}
        \path node[select vertex] at (\vertex) {${f}_{\vertex}^{(1)}$};
    
    \begin{pgfonlayer}{background}
        \foreach \source / \dest in {1/0, 3/2}
            \path[selected edge] (\source.center) -- (\dest.center);
    \end{pgfonlayer}

    	\path [draw=black, smooth ,fill=camdrk, fill opacity=0.1, very thick]
       ([xshift=-0.5em,yshift=0.5em]0.north west) -- ([xshift=0.5em,yshift=0.5em]3.north east) -- ([xshift=0.5em,yshift=-0.5em]2.south east) -- ([xshift=-1.1em,yshift=-0.5em]1.south west) -- cycle;
    
 %%-------------- level 3 ----------------------------------------------

    \foreach \pos/\name in {{(0.6875+4,0.6)/0}, {(2.875+4,0.4375)/1}}
        \node[vertex] (\name) at \pos {};
    \foreach \source/ \dest /\weight in {0/1/7}
        \path[edge] (\source) -- (\dest);

  \foreach \vertex / \fr in {0/4,1/4}
        \path node[select vertex] at (\vertex) {${f}_{\vertex}^{(2)}$};

    \begin{pgfonlayer}{background}
        \foreach \source / \dest in {0/1}
            \path[selected edge] (\source.center) -- (\dest.center);
    \end{pgfonlayer}

  \path [draw=black, smooth, fill=mygreen, fill opacity=0.1, very thick]
       ([xshift=-0.5em,yshift=0.5em]0.north west) -- ([xshift=0.5em,yshift=0.5em]1.north east) --([xshift=0.5em,yshift=-0.5em]1.south east)  -- ([xshift=-0.5em,yshift=-0.5em]0.south west) -- cycle;

    	\draw[-stealth, ultra thick,decoration={snake, pre length=0.01mm, segment length=2mm, amplitude=0.3mm, post length=1.5mm}, decorate] ([xshift=0.5em, yshift=2.5em]2.east) -- node[below, black] {} node[above] {Coarsen} ([xshift=-0.5em, yshift=-0.26em]0.west);

    %%-------------- level 4coarsest----------------------------------------------
    
        \foreach \pos/\name/\alias in {{(1.78125+7,0.51875)/0/a}}
        \node[vertex] (\name) at \pos {};
    \foreach \vertex / \fr in {0/2}
        \path node[select vertex] at (\vertex) {${f}_{\vertex}^{(3)}$};

  \path [draw=black, smooth, fill= mymauve, fill opacity=0.1, very thick]
       ([xshift=-0.5em,yshift=0.5em]0.north west) -- ([xshift=0.5em,yshift=0.5em]0.north east) -- ([xshift=0.5em,yshift=-0.5em]0.south east) -- ([xshift=-0.5em,yshift=-0.5em]0.south west)  -- cycle;
       
 	\draw[-stealth, ultra thick,decoration={snake, pre length=0.01mm, segment length=2mm, amplitude=0.3mm, post length=1.5mm}, decorate] ([xshift=0.5em]1.east) -- node[below, black] {} node[above] {Coarsen} ([xshift=-0.5em, yshift=-0.21em]0.west);

\end{tikzpicture}
\caption{An example of the tree-like hierarchical coarsening of the graph in Fig. \ref{fig:pairGraph} during the Haar transform in Eq. \eqref{eq:waveletHierarchy}.} \label{fig:HierarchicalCoarsening}
\end{figure*}
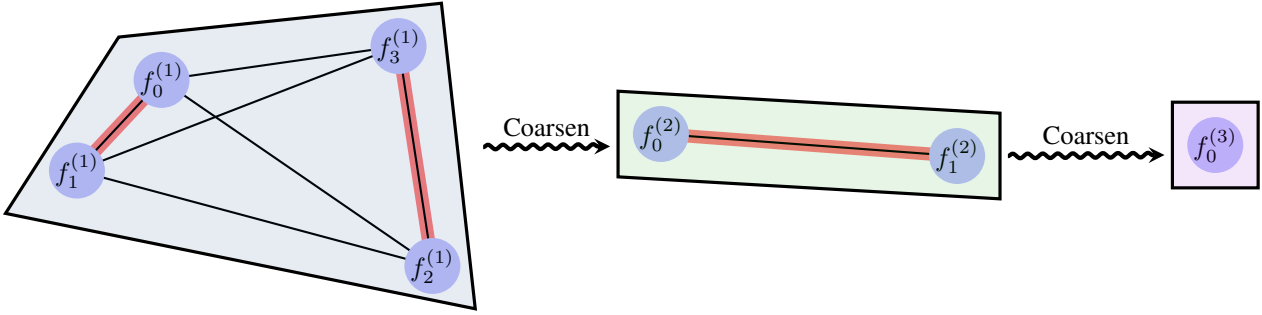

%\end{document}

\textbf{Preliminaries.}
Assume we are given an undirected graph $\cal G=({\cal V},{\cal E})$  where $\cal V$ is a set of $n$ vertices and $\cal E$ is a set of $m$ edges. Let us denote by ${\bf f}_i\in\mathbb{R}^c$ the feature vector that resides at the $i$-th node of $\cal G$ with $c$ being the number of channels. Also, we denote the degree of the $i$-th node by $d(i)$.
We define $\bf D$, the gradient matrix of the graph as follows:
\begin{equation}
    \label{eq:graphGradient}
    \textstyle ({\bf D} {\bf f})_{(i,j) \in \mathcal{E}} = {\frac{1}{\sqrt{2}}}({\bf f}_i - {\bf f}_j) ,
\end{equation}
where nodes $i$ and $j$ are connected via the $(i,j)$-th edge. This operator is a mapping from the \textit{vertex} space to the \textit{edge} space,
and can be thought of as a weighted directional derivative in the direction defined by the $(i,j)$-th edge.
Similarly to Eq. \eqref{eq:graphGradient}, we define the node-averaging operator
\begin{equation}
    \label{eq:graphAve}
    \textstyle ({\bf A} {\bf f})_{(i,j) \in \mathcal{E}} = {\frac{1}{\sqrt{2}} } ({\bf f}_i + {\bf f}_j),
\end{equation}
which also maps from the \textit{vertex} space to the \textit{edge} space. The transpose of the nodal average, i.e., ${\bf A}^\top$ is an averaging operator for the edge features.

\textbf{Pair-graph.} Given a graph $\cal G=({\cal V},{\cal E})$, we define its undirected \textit{pair-graph}
\begin{equation}
    \label{eq:pairgraph}
    {\cal G}_p=({\cal V},{{\cal E}_p}),
\end{equation}
where ${\cal E}_p\subset {\cal E}$ is a sub-set of the edges, that forms a perfect matching of the vertices. That is, we construct a graph where each vertex is paired to exactly one (different) vertex, and the degree of each vertex is exactly 1. See the example in Fig. \ref{fig:pairGraph}.  Formally, we wish to find a sub-set of the edges such that:
\begin{equation}
    \label{eq:subsetConditions}
    {\cal E}_p = \{ (i,j) \in {\cal E} \  | \ d_{p}(i) = d_{p}(j) = 1 \}
\end{equation} where $d_{p}(i)$ is the degree of the i-th node of the graph ${\cal{G}}_{p}$.
Note, that this sub-set is not unique and can be found under different criteria. In this work, we focus on the graph Haar wavelet transform and feature maps compression. Therefore, we construct ${\cal E}_p$ such that the similarity between connected vertices is maximal, subject to the constraints in Eq. \eqref{eq:subsetConditions}, to promote the sparsity of the Haar wavelet transform. This can be approximately achieved by choosing vertices that are closest in their feature norm ($\ell_2$ or $\ell_1$). Specifically, we use the Graclus \cite{dhillon2007weighted} algorithm with edge weights that are computed as $w_{ij} = \|\bff_i - \bff_j\|_2$ to efficiently obtain the pair-graph. While not optimal, this clustering algorithm is fast and parallel (in many cases, it is also used in GCNs for pooling). Following this process, if a few nodes are left with no match, we match those nodes randomly. Typically, only a small fraction of the nodes do not belong to a pair (in our experiments, under 2 $\%$), hence this has insignificant impact on the performance.

%\textbf{Orthogonal basis for graph signals.} 
\textbf{Single-level graph Haar transform.}
Given a pair-graph ${\cal{G}}_p$, we define a single level graph Haar operator as the combination of the pair-graph gradient matrix $\bf D$ and average operator $
 \bf A$: 
 \begin{equation}
     \label{eq:graphWavelet}
     \bf W = [\bf D \ \bf A].
 \end{equation}
Given node feature maps ${\bf f} \in \mathbb{R}^{n \times c}$ defined over the vertices of the graph, we compute the (single level) Haar wavelet transform of the features as follows:
 \begin{equation}
     \label{eq:signalProj}
     \bf p = \bf W f = [D\bf f \  A\bf f]
 \end{equation}
Note, that by the construction of the pair-graph, the two operators, satisfy $\bf A^\top D = 0$, since they correspond to the orthogonal basis vectors $\frac{1}{\sqrt{2}}[1,1]$, and $\frac{1}{\sqrt{2}}[1,-1]$. Also, both $\bf D^\top D$ and $\bf A^\top A$ are identity matrices. This means that the operator in Eq.
 \eqref{eq:graphWavelet} is orthogonal as well, i.e., it holds that $\bf W \bf W^{\top} = \bfI$. Thus, it can represent the feature maps $\bf f$ without loss of information. Also, since we have that the degree of each vertex is 1 in the pair-graph, it follows that the number of edges is exactly half the number of the vertices, i.e., $|{\cal{E}}_p| = \frac{|\cal{V}|}{2}$. Therefore, the gradient and averaging operators project $\bf f$ to a graph coarsened by a factor of 2.
 
\textbf{Multi-level graph Haar transform.}
The description above concludes the first Haar level, and to achieve satisfactory compression, we repeat the construction of the wavelet operator \eqref{eq:graphWavelet}, each time based on the graph averaged features $\bf A f$ to obtain a coarser representation of the original feature maps $\bff$. More precisely, denote by $\bfp^{(l)}$ the transformed features at the $l$-th graph resolution level, and by $\bff^{(l)}$ the averaged signal at the $l$-th level, both of size $\frac{n}{2^l}$ (the original input $\bff$ is equivalent to $f^{(0)}$). We define the hierarchical sequence:
\begin{eqnarray}\label{eq:waveletHierarchy}
(\bfp^{(1)},\bff^{(1)}) &=& (\bfD^{(1)}\bff, \bfA^{(1)}\bff) \nonumber \\
(\bfp^{(2)},\bff^{(2)}) &=& (\bfD^{(2)}\bff^{(1)}, \bfA^{(2)}\bff^{(1)}) \nonumber \\
&\vdots& \nonumber\\
(\bfp^{(L)},\bff^{(L)}) &=& (\bfD^{(L)}\bff^{(L-1)}, \bfA^{(L)}\bff^{(L-1)})
\end{eqnarray}
where $\bfD^{(l)},\bfA^{(l)}$ are the gradient and averaging operators for the $l$-th graph resolution, respectively. An example of a hierarchy of graphs is shown in Fig. \ref{fig:HierarchicalCoarsening}. The transformed signal is the concatenation of all the signals, including the final average:
\begin{equation}
\bfp = [\bfp^{(1)},\bfp^{(2)},\cdots,\bfp^{(L)},\bff^{(L)}] = \bfW\bff.
\end{equation}
 Note, that the total size of $\bfp$ is $n$, as $\bf f$, and since $\bfW$ is orthogonal, we have that $\|\bfp\|_2 = \|\bff\|_2$. The main feature of the Haar transform is that if the signal $\bff$ is piecewise constant or at least smooth, then $\bfp$ will be sparse, on which we elaborate next section. 

\begin{figure}
     \centering
     \begin{subfigure}[b]{0.15\textwidth}
         \centering
         \includegraphics[width=\textwidth]{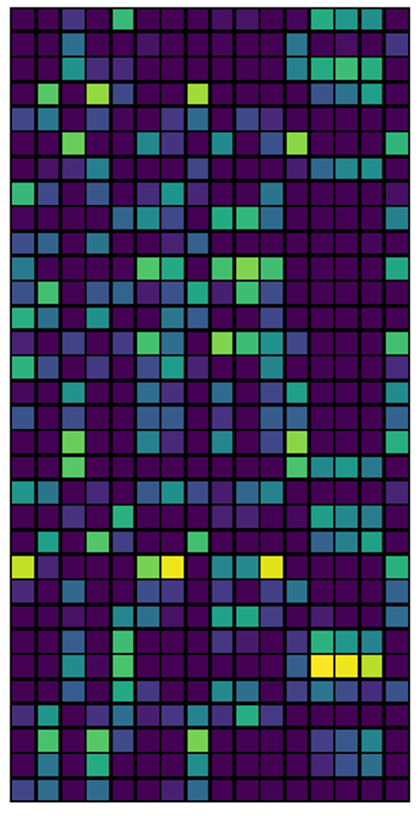}
         \caption{}
         \label{fig:inputFeatures}
     \end{subfigure}
     \hfill
     \begin{subfigure}[b]{0.15\textwidth}
         \centering
         \includegraphics[width=\textwidth]{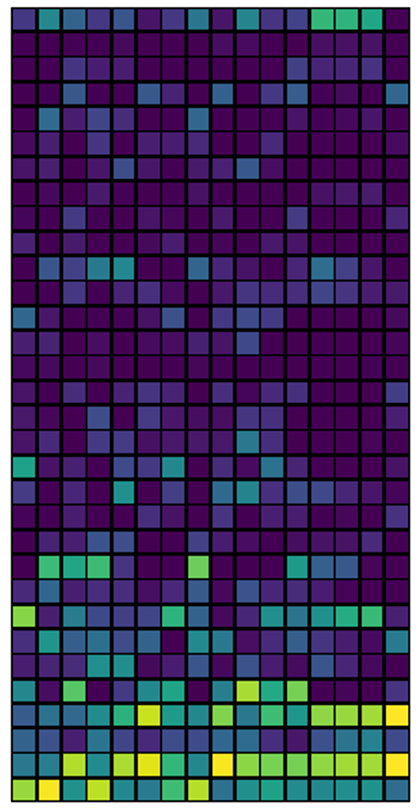}
         \caption{}
         %\caption{Wavelet transform}
         \label{fig:waveletTransform}
     \end{subfigure}
     \hfill
     \begin{subfigure}[b]{0.16\textwidth}
         \centering
         \includegraphics[width=\textwidth]{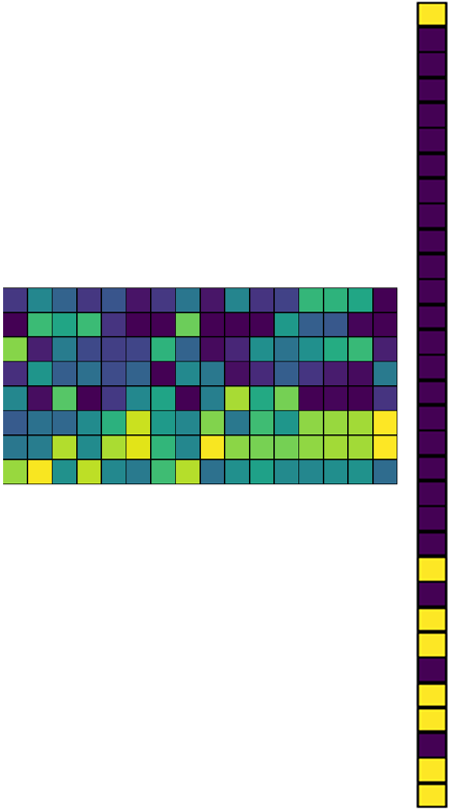}
         \caption{}
         \label{fig:compressedSignal}
     \end{subfigure}
        \caption{The flow of the Haar wavelet compression. (a) The input features---16 channels (columns) and 32 points (rows) from the second layer of DGCNN for the ModelNet40 dataset. (b) The Haar transformed graph signal (each column denote a channel). (c) The $\times 4$ compressed signal and its bitmap using joint channel sparsity. }
        \label{fig:flowfig}
\end{figure}

\subsection{Wavelet signal compression.}  Wavelet compression methods involve the lossless wavelet transformation described above, followed by a lossy shrinkage operation, which reduces the small-magnitude non-zero entries of the transformed input signal $\bfp$. Thus, we have a sparse representation of the signal in the wavelet domain, which can be transformed back to the original spatial domain via the corresponding inverse wavelet transform.

However, naively storing a multi-channel signal as sparse matrices has the overhead of maintaining the non-zero location lists, or using a bit-wise mapping per channel to each entry, which adds a bit per entry (whether it is a non-zero or not), leading to high memory costs.  Furthermore, multiplying general sparse matrices has overhead and is costly to perform in hardware. Hence, we apply the shrinkage selection jointly across all channels, such that that all channels share the same non-zero location list. Then, we gather the non-zeros of all the channels to a dense 2D tensor, which is fed into a standard $1 \times 1$ convolution operator that operates efficiently in dense arithmetic, on a significantly smaller input signal. Using the non-zero list, we apply the inverse transform after zero-filling the channels (also called ``scattering'').

Namely, given the transformed tensor $\bfp \in \mathbb{R}^{n \times c}$, we compress it by choosing a sub-set of the $n$ entries according to the feature norm of each entry. To this end, we choose $\alpha n$ entries of $\bfp$, where $ 0 < \alpha < 1$, with the largest $\ell_2$ norm, and denote the set of indices of chosen vertices by $C$. This is natural as $\bfp$ is effectively sparse, as depicted from Fig. \ref{fig:waveletTransform}. Formally, we construct the shrinkage matrix $\bfT\in\mathbb{R}^{\lceil\alpha n\rceil\times n}$, to be an operator that projects $\bfp$ onto the entries corresponding to indices in $\CC$. The compressed wavelet transform and a bit-map that indicates $\CC$ of the prior example is given in Fig. \ref{fig:compressedSignal}.

Analogously, we define the opposite operator of $\bfT$, that interpolates (zero-fills) from $\lceil\alpha n\rceil$ to $n$ entries as its transpose $\bfT^{\top}$. If indeed the signal $\bff$ is efficiently compressed by a wavelet transform (i.e., $\bfW\bff$ is sparse) then we can say that
\begin{equation}\label{eq:WaveletCompr}
\bff \approx \bfW^\top\bfT^\top\bfT\bfW\bff.
\end{equation}
 We elaborate on this claim experimentally, in Sec. \ref{subsec:validationofJointSparsity}

\subsection{Compressed wavelet transform convolution}
\label{subsec:compressed_wavelet_equiv}
The combination of the wavelet transform $\bf W$ with the compression matrix $\bfT$ in \eqref{eq:WaveletCompr}, together with a $1\times 1$ convolution matrix $ \bfK_{1\times 1}$ yields the \textit{compressed wavelet transform convolution}:
 \begin{equation}
     \label{eq:compressedWaveletTransform}
     \bff^{out} = \bfW^{\top} \bfT^{\top}  \bfK_{1\times 1} \bfT \bfW \bff^{in},
 \end{equation}
where $\bfK_{1\times 1}$ is a trainable convolution matrix. That is the core of our approach for obtaining an efficient, compressed convolution. Since $\bfT$ and $\bfW$ operate similarly on all channels, $\bfK_{1\times 1}$ does not necessarily be square (i.e., input channels need not be equal to the output channels). Here, we focus on $1\times 1$ convolutions as these are fundamental in neural networks, particularly in GCNs, where the spatial operations are done separately. Note that $\bfK_{1\times 1}$ operates on the compressed signals, after the shrinkage operator $\bfT$.

\textbf{Equivalence of convolution order.}
A $1\times 1$ convolution operates on any channel-wise input as follows: 
\begin{equation}\label{eq:Klean_g2}
\bfy^i = (\bfK_{1\times 1}\bff)^i = \sum_{j}k_{i,j}\bff^j
\end{equation}
where $\bff^j\in\mathbb{R}^n$ is the $j$-th input channel and $k_{i,j}\in\mathbb{R}$ are the weights that comprise the matrix $\bfK_{1\times 1} \in \mathbb{R}^{c_{in} \times c_{out}}$ where $c_{in}$ and $c_{out}$ are the input and output number of channels, respectively. Note, that the operation in Eq. \eqref{eq:Klean_g2} is a simple matrix multiplication in the channel space. Assume that we are applying the spatial wavelet shrinkage operator $\bfT\bfW$ that applies the same transformation and shrinkage on all channels:

\begin{equation}
\bfy^i = (\bfT\bfW\bfK_{1\times 1}\bff)^i = \bfT\bfW\sum_{j}k_{i,j}\bff^j = \sum_{j}k_{i,j}\bfT\bfW\bff^j. 
\end{equation}
It follows that the two operators commute 
\begin{equation}
    \label{eq:commute}
   \bfT\bfW	\bfK_{1\times 1} = \bfK_{1\times 1}\bfT\bfW.
\end{equation}
In fact, the result above holds for any arbitrary spatial operation, such as node gradient and Laplacian, and also our inverse wavelet shrinkage transform $\bfW^\top\bfT^\top$.
Thus, the operation in Eq. \eqref{eq:compressedWaveletTransform} is equivalent to a wavelet compression (Eq. \eqref{eq:WaveletCompr}) \emph{after} a $1\times1$ convolution  
\begin{eqnarray}\label{eq:convAfter}
\bff^{out} &=& \bfW^\top\bfT^\top\bfT\bfW\bfK_{1\times 1} \bff^{in} \nonumber \\ &=& \bfK_{1\times 1}\bfW^\top\bfT^\top\bfT\bfW \bff^{in}.
\end{eqnarray}
The main difference between \eqref{eq:compressedWaveletTransform} and \eqref{eq:convAfter} is the computational cost: in \eqref{eq:compressedWaveletTransform} the convolution operates on the \emph{dense} compressed wavelet transform. We state the following lemma for a GCN layer.  

\begin{lemma}
\label{lemma:orderequiv}
Given node features $\bff$, it holds that a $1\times 1$ convolution layer with an activation function $\sigma$ satisfies:
%$$ \sigma(KW^{-1}(T^{T}TW(f))) = \sigma(W^{-1}(T^{T} KTW(f))) $$
\begin{equation} \sigma( \bfK_{1\times 1}\bfW^\top\bfT^\top\bfT\bfW \bff) = 
\sigma( \bfW^\top\bfT^\top \bfK_{1\times 1} \bfT\bfW \bff)
\end{equation}
and if no compression is used ($\bfT=\bfI$) then 
\begin{equation} \sigma( \bfK_{1\times 1} \bff) = 
\sigma( \bfW^\top\bfK_{1\times 1} \bfW \bff)
\end{equation}
\end{lemma}
The proof follows immediately from Eq. \eqref{eq:compressedWaveletTransform} and \eqref{eq:convAfter}, and the fact that $\bfW$ is orthogonal.

Thus, we can perform the convolution in the wavelet domain with dense $\lceil\alpha n\rceil$ elements. As long as the wavelet compression is effective (i.e., $\bfW \bff$ is sparse), we obtain a similar result as applying the convolution on the dense feature tensor with $n$ elements. $\alpha$ is typically chosen to be $\frac{1}{4}$ or $\frac{1}{8}$. Note that in other works mentioned earlier, it is the spatial separable convolution that is applied in the wavelet domain, while the $1\times 1$ convolution is applied in the original domain. Also, note that the result above holds regardless of the entries in $\bfK_{1\times 1}$. Hence, we can use our transform to compress the feature maps of a standard network with no wavelets, given pre-trained weights. This scenario is especially important in cases where we have no access to the data and wish to compress the network without re-training (because of privacy issues, for example).

In addition, we propose an alternative scheme that includes less wavelet transformations by combining two subsequent convolution layers after applying one wavelet transform. That is, we stack multiple convolutions in the wavelet domain, instead of a single $1\times1$ convolution. For instance, an application of two convolution reads the following:
\begin{equation}
\label{eq:v2Conv}
\sigma( \bfW^\top\bfT^\top \bfK_{1\times 1}^{2}\sigma(\bfK_{1\times 1}^{1} \bfT\bfW \bff))
\end{equation}
where $\bfK_{1\times 1}^{1}$ and $\bfK_{1\times 1}^{2}$ are different trainable weights. Note that this formulation does not satisfy lemma \ref{lemma:orderequiv}, hence it requires full training. 

\subsection{Employing graph wavelet compression in GCNs} \label{wavelet_conv}
In this work, we focus on both geometric and non-geometric tasks, which are typically treated with different networks. In particular, we focus on two popular networks: DGCNN \cite{wang2018dynamic} and GCNII \cite{chen20simple}, which we compress to various degrees to demonstrate the effectiveness of our approach.

%%% UP TO HERE

\textbf{Wavelet compressed DGCNN.}
Given features tensor $\bff$ defined on the vertices of the graph $\mathcal{G} = ({\cal V}, {\cal E})$, the edge convolution \cite{wang2018dynamic} operation is given by 
\begin{equation}
    \label{eq:edgeconv}
    \bff^{(l+1)}_{i} =  \underset{(i,j) \in {\cal{E}}}{\square} \sigma( \bfK_{1\times1} [\bff^{(l)}_{i}, \bff^{(l)}_{i} - \bff^{(l)}_{j}] ),
\end{equation}
where $\bfK_{1\times1}\in\mathbb{R}^{c_{out} \times 2\cdot c_{in}}$ is the convolution matrix, $\square$ is a symmetric aggregation operator such as $\max$ or averaging.
Alternatively, we could replace the order of the spatial and convolution operations as follows \cite{li2021towards}: 
\begin{equation}
    \label{eq:edgeconvCheap}
    \bff^{(l+1)}_{i} =  \underset{(i,j) \in {\cal{E}}}{\square} \sigma(  [\bfy^{(l)}_{i} +  \bft^{(l)}_{i} - \bft^{(l)}_{j}] ),
\end{equation}
where $\bfy^{(l)} = \bfK_{1\times1}^1\bff^{(l)}$, and $\bft^{(l)} = \bfK_{1\times1}^2\bff^{(l)}$, both of size $c_{out} \times c_{in}$, and hence have the same number of parameters as in Eq. \eqref{eq:edgeconv}. Both Eq. \eqref{eq:edgeconv}
and \eqref{eq:edgeconvCheap} are equivalent, and the observation above significantly reduces the computational cost (a factor of $\frac{1}{k}$ where $k$ is the number of neighbors per node). Using Lemma \ref{lemma:orderequiv}, we simply compress the $1\times1$ convolutions:
\begin{equation*}
\bfy^{(l)} = \bfW^\top\bfT^\top\bfK_{1\times1}^1\bfT\bfW\bff^{(l)},\quad\bft^{(l)} = \bfW^\top\bfT^\top\bfK_{1\times1}^2\bfT\bfW\bff^{(l)}.
\end{equation*}

\textbf{Wavelet compressed GCNII.} A GCNII layer is defined as follows\footnote{In \cite{chen20simple}, the convolution is written to the right-hand side of the signal to operate on the channel (column) space of $\bff$: $\bff^{(l+1)} = \sigma(\bfS^{(l)} (\bff^{(l)})\bfK^{(l)}_{gcnii})$ . Here we place it as a convolution operator operator after $\bfS$ so it is consistent with the other notation in this paper. Both writings are equivalent.} \cite{chen20simple}:
\begin{equation}
    \label{eq:gcnii}
    \bff^{(l+1)} = \sigma(  \bfK^{(l)}_{gcnii} \bfS^{(l)} (\bff^{(l)})),
\end{equation}
where $\bfS^{(l)} (\cdot)$ is the spatial operation 
\begin{equation}
\bfS^{(l)} (\bff^{(l)}) =  ( (1-\alpha_{l}) \tilde{\bfP}\bff^{(l)} + \alpha_{l}\bff^{(0)}), 
\end{equation}
where $\tilde\bfP = \bfI - \tilde\bfL$ and $\tilde\bfL$ is the normalized graph Laplacian for the graph $\mathcal{G}$ with one additional self loop for each node. The addition of the initial feature helps preventing the over-smoothing phenomenon in GCNs. The channel mixture of GCNII, which is the learnt part, is given by $\bfK^{(l)}_{gcnii}$:
\begin{equation} 
\bfK^{(l)}_{gcnii} =  ( (1-\beta_{l})\bfI_{n} + \beta_{l}\bfK_{1\times1}^{(l)}  ), 
\end{equation}
where $\bfK_{1\times1}^{(l)}$ is a trainable $1\times 1$ convolution matrix. The wavelet compressed version of GCNII is written as follows:
\begin{equation}
    \label{eq:waveletGCNII}
        \bff^{(l+1)} = \sigma( \bfW^{\top}\bfT^\top\bfK^{(l)}_{gcnii}\bfT\bfW\bfS^{(l)} \bff^{(l)})
\end{equation}
$\alpha_l$ and $\beta_l$ are hyper-parameters as in \cite{chen20simple}.

\subsection{Quantization in GCNs}
We apply quantization to several components of GCNs discussed above. We quantize the weights of all layers in the GCN to a specified precision (typically 1, 2, 4, or 8 bits). Following a common practice in quantized neural networks \cite{LSQ,li2019additive}, the first and last layers are given a high bit rate -- 32 bits. That is to prevent degradation of the input or final output, as quantizing these layers often leads to a large loss in performance, while providing minimal gains in terms of the compression. The input activations of the spatial operations (wavelet, edge convolution or the graph Laplacian) and MLPs are also quantized, as well as the input activations of the wavelet transformation. The channel mixing MLPs use quantized weights. We quantize the network as is commonly done in CNNs, using the same bit rates to all the hidden layers and the inner convolutions. Our quantization is uniform, and uses the same bit-rate for all graph nodes for hardware efficiency. Obviously, making the quantization more sophisticated will improve our method, but may be less efficient on hardware. This is a trade-off that can be adapted given the situation.

%-------------------------------------------------------------------------------------------------------------

\section{Experiments}

In this section we demonstrate the efficiency of our wavelet approach for feature maps compression. We start with a basic experiment to quantify the performance of wavelets in feature-maps compression. Following that, we carry various experiments both on geometric and non-geometric datasets, for different tasks, such as node-classification, point cloud classification, and point cloud segmentation. Our main goal is to demonstrate that the Haar wavelet compression is significantly better than aggressive feature quantization, and since the operations are carried using compact dense vectors, this is also efficient in hardware, i.e., we do not use sparse arithmetic in the convolutions.

The comparison is done using the primary performance metrics of each benchmark (e.g., mean square error (MSE), accuracy, mean intersection over union (mIoU)), as well as the activations compression ratios of the convolution operations, which is the most memory intensive operation in GCNs.

Unless stated otherwise, in all the experiments we apply 8-bit weight quantization to all the networks. Furthermore, we apply a wide array of activations quantization ratios, except the first and last $1\times 1$ convolutions, which are kept at the standard 32-bit precision. Our code is implemented using PyTorch and conducted on an Nvidia Titan RTX GPU.

\subsection{Validation of joint sparsity}
\label{subsec:validationofJointSparsity}
\begin{figure}
    \centering
    \includegraphics[  width=0.5\textwidth]{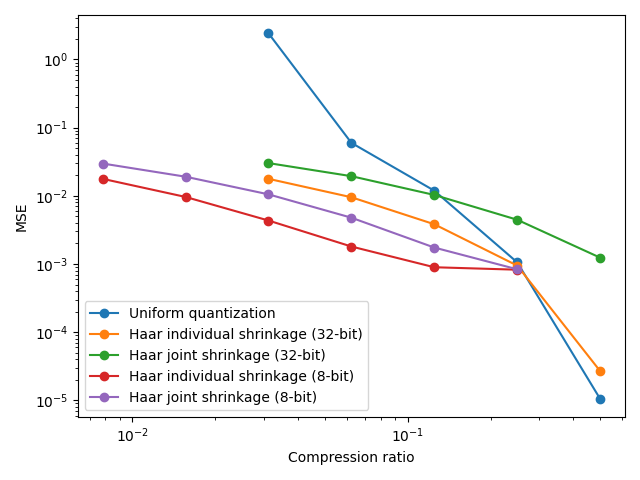}
    \caption{
    The MSE of $\times 2^q$ compression ratios (for $q=1,...,7$) and the schemes: the uniform quantization, compressed Haar transform using individual and joint channel shrinkage, and compressed and 8-bit quantized Haar transform using individual and joint channel shrinkage (used extensively in our networks). It is evident that the joint shrinkage is less accurate than the individual shrinkage, but it leads to a much more favorable implementation and computational efficiency.  
    }
    \label{fig:joinySparsityMSE}
\end{figure}

In our first experiment we demonstrate the ability of the joint-channels Haar wavelet transform to compress graph features that are taken from a network trained to classify the  ModelNet-40 dataset \cite{wu20153d}. That is, we extract learnt feature maps $\bff$, and measure the mean square-error (MSE) between $\bff$ and its compressed wavelet transform
\begin{equation}
\frac{1}{n\cdot c_{in}}\|\bff - \bfW^\top\bfT^\top\bfT\bfW\bff\|_2^2,
\end{equation}
for $\bfT$ that is applied with various compression percentages.
Fig. \ref{fig:joinySparsityMSE} shows that indeed our method is superior to standard (uniform) quantization of feature maps, and produces similar performance compared to an individual compression per channel, which while accuracy-wise is preferred, it has a tremendous cost. That is, such a compression would require storing the indices of chosen elements per channel. For typical networks with dozens and hundreds of channels, this would defeat the purpose of the feature-maps compression. More importantly, per channel compression does not satisfy the property in Lemma \ref{lemma:orderequiv}, which we rely on for efficiency (compressed dense $1\times1$ convolutions). The approach of joint channel compression is suitable both accuracy and efficiency wise.

%-------------------------------------------------------------------------------------------------------------
\iffalse
\subsection{Datasets and settings}
To compare the effectiveness of our approach with other related methods, we apply our wavelet convolution and wavelet-quantization compression scheme to a variety of GCNs, across several benchmarks. These benchmarks include semi-supervised node classification section \ref{section_semi_supervised}, shape classification \ref{section_shape_classification} and Semantic and-part segmentation of point clouds  \ref{section_semantic_segmentation}. 

The comparison is done using the primary performance metrics of each benchmark (Accuracy, mIoU, etc.), as well as the activations compression ratios of the spatial convolution operations, which are the most memory intensive operation in GCNs.

In all experiments which involve quantized GCNs, we apply 8-bit weight quantization to all layers. Furthermore, in these experiments, we apply activations quantization (typically 1, 2, 4 or 8-bits) to all layers, except the first and last $1\times 1$ convolutions, which are set to the standard 32-bit precision.
\fi

%-------------------------------------------------------------------------------------------------------------

\subsection{Semi-supervised node classification} \label{section_semi_supervised}

\begin{table}[]
    \centering
    \begin{tabular}{cccccc}
    Benchmark & Classes & Label rate & Nodes & Edges & Features \\
    \toprule
    Cora & 7 &  0.052 & 2708 & 5429 & 1433 \\
    Citeseer & 6 & 0.036 & 3327 & 4732 & 3703 \\
    Pubmed & 3 & 0.003 & 19717 & 44338 & 500\\
    \bottomrule
    \end{tabular}
    \caption{Statistics of semi-supervised benchmarks}
    \label{tab:semisupervised_statistics}
\end{table}

\begin{table}[]
    \centering
    \begin{tabular}{ccccc}
    Model & Cora & Citeseer & Pubmed \\
    \toprule

        DeepWalk \cite{deepWalk} & 67.2 & 43.2 & 65.3 \\
        Incep (4 layers) \cite{Rong2020DropEdge:} & 77.6 & 69.3 & 77.7 \\
        ChebNet \cite{defferrard2016convolutional} & 81.2 & 69.8 & 74.4 \\
        GCN\cite{kipf2016semi} & 81.5 &   70.8 & 79.0 \\
        JKNet (4 layers) \cite{jknet} & 80.2 & 68.7 & 78.2 \\
        GCNII \cite{chen20simple} & 82.2 & 68.2 & 78.2  \\
        HANet\cite{fasthaarGCN} &  81.9 & 70.1   & 79.3 \\
        GWNN\cite{GWNN} &  82.8 & 71.7   & 79.1 \\
        GAT\cite{GAT} &  83.0 & 72.3   & 79.0 \\
        NGCN\cite{NGCN} &  83.0 & 72.2   & 79.5 \\
        DualGCN\cite{DualGCN} &  83.5 & 72.4   & 79.3 \\
        HGCN\cite{HGCN} &  84.2 & 72.9   & 79.1 \\
        g-U-Nets\cite{g-U-Nets} &  84.4 & 73.2   & 79.6 \\
        DeepGWC\cite{DeepGWC} &  84.8 & 72.6   & 80.4 \\ 
        M-GWNN\cite{M-GWNN} &  84.6 & 72.6   & 80.4 \\
   \midrule
        WGCN (ours)  & $83.4$ & $71.3$ & $79.9$ \\
       WGCNII (ours)  & $84.9$ & $74.3$ & $80.6$ \\
       \bottomrule
    \end{tabular}
    \caption{Accuracy on semi-supervised node classification. 
    }
    \label{tab:baseline_semisupervised}
\end{table}

\begin{table}[]
    \centering
    \begin{tabular}{ccccccc}
    Model & W/A   & Wav.   & Total-act.  & Cora &  Cite. & Pub.   \\
     &  bits &  comp.  &  comp. &   & & \\
    \toprule
    
        GCN
        %& 32 & -- & $\times$ 1 & 81.5& 70.8 & 79.0   \\
        & 8/8 & -- & $\times$ 4 &  82.5   & 69.5    & 78.7 \\
        \cite{kipf2016semi}& 8/4 & -- & $\times$ 8 & 39.1 & 26.1    & 40.7 \\
        & 8/2 & --  & $\times$ 16 & 20.7    &  22.4 & 40.1  \\
        & 8/1 & --  & $\times$ 32 & 20.2  & 20.3 & 33.0 \\
        
        \midrule 
        GCNII
        & 8/8 & -- & $\times$ 4 & 80.9 & 69.8  & 80.0  \\
        \cite{chen20simple}& 8/4 & -- & $\times$ 8 & 31.9 & 24.7  & 41.3  \\
         & 8/2 & --  & $\times$ 16 & 21.1 & 18.3   & 40.7 \\
        & 8/1 & --  & $\times$ 32 & 20.0 & 18.0  & 36.1 \\

        \midrule
        QAT-GCN

        & 8/8 & -- & $\times$ 4 & $81.0$ & $71.3$  & -  \\
        \cite{tailor2021degree}& 4/4 & -- & $\times$ 8 & $77.2$ & $64.1$  & - \\

        \midrule
        nQAT-GCN
        & 8/8 & -- & $\times$ 4 & $81.0 $ & $70.7 $  & -  \\
        \cite{tailor2021degree}& 4/4 & -- & $\times$ 8 & $78.1 $ & $65.8$  & - \\

        \midrule
        DQ-GCN

        & 8/8 & -- & $\times$ 4 & $81.7$ & $71.0$  & -  \\
        \cite{tailor2021degree}& 4/4 & -- & $\times$ 8 & $78.3$ & $66.9$  & - \\

       \midrule 
        SGQ-GAT
        & 32/4 & -- & $\times$ 8 & 81.1 &-  & -  \\
         (uniform) \cite{SGQuant} \\ %& W4A4 & -- & $\times$ 8 & $78.3 \pm 1.7$ & $66.9 \pm 2.4$  & - \\
         \midrule
         SGQ-GCN 
        & 32/1.2 & -- & $\times$ 26.6 & 81.7 & 71.5  & 80.3  \\
        (per-node) \cite{SGQuant} \\ %& W32A4 & -- & $\times$ 8 & $78.3 \pm 1.7$ & $66.9 \pm 2.4$  & - \\
        
        \midrule
                WGCN
        & 8/8 & $\times$  1 & $\times$ 4 & 83.5  & 71.2  &    80.0  \\
       (ours) & 8/8 & $\times$ 2 & $\times$ 8 & 80.4   & 70.2   & 80.2   \\
        %& 4 & $\times$ 1 & $\times$ 8 &      \\
        & 8/8 & $\times$ 4 & $\times$ 16 & 78.1 & 70.0  &  79.5    \\
        %& 4 & $\times$ 2 & $\times$ 16 &      \\
        %& 2 & $\times$ 1 & $\times$ 16 &      \\
        & 8/8 & $\times$ 8 & $\times$ 32 & 74.2  & 63.8 &  77.6  \\
    
        %& 4 & $\times$ 4 & $\times$ 32 &      \\
        %& 2 & $\times$ 2 & $\times$ 32 &      \\
        %& 1 & $\times$ 1 & $\times$ 32 &      \\
        %& 4 & $\times$ 8 & $\times$ 64 &      \\
        %& 2 & $\times$ 4 & $\times$ 64 &      \\
        %& 1 & $\times$ 2 & $\times$ 64 &      \\
        %& 2 & $\times$ 8 & $\times$ 128 &     \\
        %& 1 & $\times$ 4 & $\times$ 128 &      \\
        \midrule
        WGCNII
        & 8/8 & $\times$  1 & $\times$ 4 & 84.5 & 74.1 & 80.4     \\
       (ours) & 8/8 & $\times$ 2 & $\times$ 8 & 84.9  & 73.7  & 80.7  \\
        %& 4 & $\times$ 1 & $\times$ 8 &      \\
        & 8/8 & $\times$ 4 & $\times$ 16 & 83.2 & 73.3 & 79.1     \\
        %& 4 & $\times$ 2 & $\times$ 16 &      \\
        %& 2 & $\times$ 1 & $\times$ 16 &      \\
        & 8/8 & $\times$ 8 & $\times$ 32 & 82.1 & 71.8 & 77.3    \\
        %& 4 & $\times$ 4 & $\times$ 32 &      \\
        %& 2 & $\times$ 2 & $\times$ 32 &      \\
        %& 1 & $\times$ 1 & $\times$ 32 &      \\
        %& 4 & $\times$ 8 & $\times$ 64 &      \\
        %& 2 & $\times$ 4 & $\times$ 64 &      \\
        %& 1 & $\times$ 2 & $\times$ 64 &      \\
        %& 2 & $\times$ 8 & $\times$ 128 &     \\
        %& 1 & $\times$ 4 & $\times$ 128 &      \\
    \bottomrule
        
    \end{tabular}
    \caption{Accuracy ($\%$) on semi-supervised node classification. W/A indicates the weights and activations, respectively.}
    \label{tab:quantized_semisupervised}
\end{table}

In this section we evaluate our method on three citation network datasets: Cora, Citeseer and Pubmed \cite{sen2008collective}. The statistics of the datasets can be found in Table \ref{tab:semisupervised_statistics}. On each dataset we use the standard training/validation/testing split as in \cite{yang2016revisiting}, with 20 nodes per class for training, 500 validation nodes and 1,000 testing nodes and follow the training scheme of \cite{chen20simple},
where we adopt the GCN\cite{kipf2016semi} and GCNII \cite{chen20simple} architectures, and replace each convolution kernel with our analogous wavelet convolution.  
Those models are denoted as WGCN and WGCNII, respectively.
We set the learning rate to $0.01$, with a weight decay of $5\cdot 10^{-4}$. The dropout is $0.6\ , 0.7 \ , 0.5$ on Cora/Citeseer/Pubmed respectively. As in \cite{chen20simple}, we set the hyper parameters of WGCNII to a fixed $\alpha=0.1$ for all layers, and  $\beta_l = log(\frac{\lambda}{l}+1)$ with $\lambda=0.1$.
To allow a fair comparison with various popular methods like GCN, GCNII, GAT as well as other wavelet based GCNs: HANet\cite{fasthaarGCN}, DeepGWC\cite{DeepGWC} and M-GWNN\cite{M-GWNN}, we use a two-layer network, and do not perform quantization or compression. For JKNet \cite{jknet} and Incep \cite{Rong2020DropEdge:} we compare with the minimal number of layers that are reported.
The results of our wavelet convolution can be seen in Tab.\ref{tab:baseline_semisupervised}, where we read a higher or same accuracy across all datasets, compared to the considered methods.

Furthermore, we experiment with different compression ratios of the wavelet transform and activations quantization. We compare our method to several recent works such as QAT \cite{tailor2021degree} and SGQuant \cite{SGQuant}. We note that such methods apply component and layer wise as well as topology-aware quantization, which are considerably more sophisticated than our uniform (layer, channel and node wise) quantization. Such algorithmic upgrades can have significant costs on low computational resource, edge devices. We apply our uniform quantization to the baseline methods of GCN and GCNII (followed by a re-run evaluation) and their counterparts WGCN and WGCNII. The results are reported in Tab. \ref{tab:quantized_semisupervised}, where two contributions are portrayed. First, we see that compared to the GCN and GCNII, their wavelet variants WGCN and WGCNII (ours) yield significantly higher accuracy, even at larger compression rates. In addition, we see that compared to quantization designated methods, we obtain comparable or higher accuracy per compression ratio.

%-------------------------------------------------------------------------------------------------------------

\subsection{Fully-supervised node classification}
We evaluate our method on fully supervised datasets Cora, Cornell, Texas and Wisconsin \cite{Pei2020Geom-GCN:}, with the same train/validation/test splits of $60 \%, 20\%, 20\%$ and 10 random splits from \cite{Pei2020Geom-GCN:}. For each dataset, we use the hyper parameters reported in \cite{chen20simple}. We compare two-layer GCNII and WGCNII, where we see similar results when the activations compression is under $\times 16$,  and significantly better accuracy as the compression increases compared to more a aggressive quantization---see Tab. \ref{tab:quantized_FullySupervised}.

\begin{table}[]
    \centering
    \begin{tabular}{cccccc}
    Model   & Total-act.  & Cora &  Cornell &  Texas & Wisconsin   \\
     &   comp. &   & & \\
    \toprule
    
        GCNII
         & $\times$ 4 &  88.0 & 86.2 & 83.51  & 89.8  \\
        \cite{chen20simple} & $\times$ 8 & 87.8  & 85.4 & 85.1 & 89.6   \\
         & $\times$ 16 & 30.2 & 58.9 & 70.2 & 71.6  \\
         & $\times$ 32 & 28.9 & 57.8 & 58.9 & 48.0 \\

        \midrule
        WGCNII
        & $\times$ 4 & 88.9 & 86.7 &  84.6 & 89.5     \\
       (ours) & $\times$ 8 & 89.0 & 88.6  & 85.1 & 89.0 \\

        & $\times$ 16 & 87.9  & 87.0  & 84.0 & 88.1  \\

         & $\times$ 32 & 86.1 & 85.9 & 83.3 & 86.8    \\
    \bottomrule
        
    \end{tabular}
    \caption{Accuracy ($\%$) on fully-supervised node classification. Total-activations compression follows the same settings as in Tab. \ref{tab:quantized_semisupervised}. 
    }
    \label{tab:quantized_FullySupervised}
\end{table}

%-------------------------------------------------------------------------------------------------------------
\subsection{Shape classification} \label{section_shape_classification}

\begin{table}[]
    \centering
    \begin{tabular}{ccccc}
    Model & Act.  & Wav.   & Total-act.  & Overall   \\
     &  bits &  comp.  &  comp. & Acc. $\%$   \\
        \toprule
        DGCNN
         & 8 & -- &  $\times$ 4 &    92.2   \\
        & 4 & -- &  $\times$ 8 &    89.7    \\
        & 2 & --  &  $\times$ 16 &   87.1    \\
        & 1 & --  &  $\times$ 32 &   80.7     \\
        
        \midrule
        WDGCNN\textsubscript{V \textsubscript{1}}
        & 8 & $\times$  1 & $\times$ 4 & 92.2      \\
        & 8 & $\times$ 2 & $\times$ 8 &   92.4  \\
        %& 4 & $\times$ 1 & $\times$ 8 &   \\
        & 8 & $\times$ 4 & $\times$ 16 &   92.0   \\
        %& 4 & $\times$ 2 & $\times$ 16 &      \\
        %& 2 & $\times$ 1 & $\times$ 16 &      \\
        & 8 & $\times$ 8 & $\times$ 32 & 91.1     \\
        & 8 & $\times$ 16 & $\times$ 64 &  90.6    \\
        %& 4 & $\times$ 4 & $\times$ 32 &      \\
        %& 2 & $\times$ 2 & $\times$ 32 &      \\
        %& 1 & $\times$ 1 & $\times$ 32 &      \\
        %& 4 & $\times$ 8 & $\times$ 64 &      \\
        %& 2 & $\times$ 4 & $\times$ 64 &      \\
        %& 1 & $\times$ 2 & $\times$ 64 &      \\
        %& 2 & $\times$ 8 & $\times$ 128 &     \\
        %& 1 & $\times$ 4 & $\times$ 128 &     \\
        \midrule
        WDGCNN\textsubscript{V \textsubscript{2}}
        & 8 & $\times$  1 & $\times$ 4 & 91.9     \\
        & 8 & $\times$ 2 & $\times$ 8 & 91.4    \\
        & 8 & $\times$ 4 & $\times$ 16 &  90.7    \\
        & 8 & $\times$ 8 & $\times$ 32 &   89.4   \\
        & 8 & $\times$ 16 & $\times$ 64 &  88.5    \\
\bottomrule
        
    \end{tabular}
    \caption{Shape classification accuracy on ModelNet40.}
    \label{tab:quantized_modelenet_results}
\end{table}

We demonstrate our method for the 3D shape classification benchmark ModelNet40 \cite{wu20153d}. The dataset consists of 12,311 CAD meshes, across 40 categories, with 9,843 and 2,468 samples in the train and test sets, respectively. 
We randomly sample 1,024 points from each mesh, and normalize the points to the unit cube. We follow \cite{wang2018dynamic} and construct a graph from each point cloud, using the kNN algorithm with $k = 20$.

Our network adopts the popular DGCNN architecture \cite{wang2018dynamic}, in which each DGCNN block is replaced with the equivalent wavelet block from Eq. \eqref{eq:edgeconvCheap}, and is denoted by WDGCNN\textsubscript{V\textsubscript{1}}. Also, we examine the performance with the alternative convolution from Eq. \eqref{eq:v2Conv}, denoted by WDGCNN\textsubscript{V\textsubscript{2}}.
Our training scheme is as in \cite{wang2018dynamic}, where we use the Adam \cite{kingma2014adam} optimizer with a learning rate of 0.01 and a a step scheduler with a decrease factor of 0.5 every 20 epochs, for 250 epochs. All our evaluations are re-runs. The results are summarized in Table \ref{tab:quantized_modelenet_results}, where it is noted that our method outperforms activations quantizations, revealing a healthy margin when aggressive quantization (i.e., 2 or 1 bits) are applied.

%-------------------------------------------------------------------------------------------------------------

\begin{table}
    \centering
    \begin{tabular}{ccccc}
    Model & Act.  & Wav.   & Total-act.  & Instance    \\
     &  bits &  comp.  &  comp. & mIoU.   \\
        \toprule
        DGCNN
        & 8 & -- &  $\times$ 4 &  84.1     \\
        & 4 & -- &  $\times$ 8 &    80.7    \\
        & 2 & --  &  $\times$ 16 &   75.6 \\
        & 1 & --  &  $\times$ 32 &   38.1     \\
        \midrule
        WDGCNN\textsubscript{V \textsubscript{1}}
        & 8 & $\times$  1 & $\times$ 4 & 84.3      \\
        & 8 & $\times$ 2 & $\times$ 8 &  83.7   \\
        %& 4 & $\times$ 1 & $\times$ 8 &   \\
        & 8 & $\times$ 4 & $\times$ 16 &   82.1    \\
        %& 4 & $\times$ 2 & $\times$ 16 &      \\
        %& 2 & $\times$ 1 & $\times$ 16 &      \\
        & 8 & $\times$ 8 & $\times$ 32 &  81.5   \\
        
        & 8 & $\times$ 16 & $\times$ 64 & 78.1  \\ 
        %& 4 & $\times$ 4 & $\times$ 32 &      \\
        %& 2 & $\times$ 2 & $\times$ 32 &      \\
        %& 1 & $\times$ 1 & $\times$ 32 &      \\
        %& 4 & $\times$ 8 & $\times$ 64 &      \\
        %& 2 & $\times$ 4 & $\times$ 64 &      \\
        %& 1 & $\times$ 2 & $\times$ 64 &      \\
        %& 2 & $\times$ 8 & $\times$ 128 &     \\
        %& 1 & $\times$ 4 & $\times$ 128 &     \\
        
        \midrule
        WDGCNN\textsubscript{V \textsubscript{2}}
        & 8 & $\times$  1 & $\times$ 4 &  84.1     \\
        & 8 & $\times$ 2 & $\times$ 8 &   83.9   \\
        %& 4 & $\times$ 1 & $\times$ 8 &   \\
        & 8 & $\times$ 4 & $\times$ 16 &  83.0     \\
        %& 4 & $\times$ 2 & $\times$ 16 &      \\
        %& 2 & $\times$ 1 & $\times$ 16 &      \\
        & 8 & $\times$ 8 & $\times$ 32 &    80.6   \\
        & 8 & $\times$ 16 & $\times$ 64 &   78.5   \\
        %& 4 & $\times$ 4 & $\times$ 32 &      \\
        %& 2 & $\times$ 2 & $\times$ 32 &      \\
        %& 1 & $\times$ 1 & $\times$ 32 &      \\
        %& 4 & $\times$ 8 & $\times$ 64 &      \\
        %& 2 & $\times$ 4 & $\times$ 64 &      \\
        %& 1 & $\times$ 2 & $\times$ 64 &      \\
        %& 2 & $\times$ 8 & $\times$ 128 &     \\
        %& 1 & $\times$ 4 & $\times$ 128 &     \\
        \bottomrule
    \end{tabular}
    \caption{Part segmentation on ShapeNet.}
    \label{tab:quantized_shapenet_results}
\end{table}

\subsection{Semantic and part segmentation} \label{section_semantic_segmentation}

In this section we evaluate our approach on two different segmentation datasets: the Shapenet part segmentation \cite{chang2015shapenet} and
Stanford Large-Scale 3D Indoor Spaces Dataset (S3DIS) \cite{armeni20163d}. 

The Shapenet part segmentation \cite{chang2015shapenet} dataset includes 16,881 3D point clouds, across 16 shape categories, with a total of 50 part annotation classes (each shape contains 2-6 parts). We sample 2,048 points from each shape, where the goal is to correctly classify these part annotation per point. The train, validation and test sets are split according to \cite{chang2015shapenet}.

The Stanford Large-Scale 3D Indoor Spaces Dataset (S3DIS) \cite{armeni20163d}, is another semantic segmentation benchmark. The dataset includes 3D scans of 272 rooms from 6 different areas. Each point in these 3D scans is annotated with one of 13 semantic classes, and we wish to classify each point correctly. We adopt the pre-processing steps of splitting each room into 1m × 1m blocks and sample 4,096 points from the 3D scan. Each point is represented by a 9D vector (XYZ, RGB, normalized spatial
coordinates). The train, validation and test splits are the same as in \cite{qi2017pointnet}, and follows the 6-fold evaluation protocol\cite{armeni20163d}.

For both datasets, we employ the same architecture, training and testing schemes as in DGCNN, only replacing each edge-conv block (as in Eq. \eqref{eq:edgeconv}) with ours from Eq. \eqref{eq:edgeconvCheap}.
For constructing the graph we use kNN with $k=20,40$ for Shapenet and S3DIS, respectively.
Our results on Shapenet part-segmentation are provided in Tab. \ref{tab:quantized_shapenet_results}, and for S3DIS in Tab. \ref{tab:quantized_S3DIS_results}, and for both datasets, re-run of DGCNN is performed. We notice similar or slightly better performance when the compression ratio is under $\times8$. More significantly, we see that for extreme compression of over $\times16$, our WDGCNN\textsubscript{V\textsubscript{1}} maintains high accuracy, compared to DGCNN where a major degradation of the accuracy takes place. Also, for ShapeNet, we report the accuracy of WDGCNN\textsubscript{V\textsubscript{2}}, where we see similar performance to WDGCNN\textsubscript{V\textsubscript{1}}, with less transformations.

\begin{table}
    \centering
    \begin{tabular}{ccccc}
    Model & Act.  & Wav.   & Total-cct.  & mIoU.   \\
     &  bits &  comp.  &  comp. &   \\
        \toprule
        DGCNN
        & 8 & -- &  $\times$ 4 & 56.5      \\
        & 4 & -- &  $\times$ 8 & 55.8        \\
        & 2 & --  &  $\times$ 16 &  53.1      \\
        & 1 & --  &  $\times$ 32 &  32.1      \\

        \midrule
        WDGCNN\textsubscript{V \textsubscript{1}}
        %& 8 & $\times$  1 & $\times$ 4 &      \\
        & 8 & $\times$ 2 & $\times$ 8 & 56.9    \\
        & 8 & $\times$ 4 & $\times$ 16 & 56.0      \\
        & 8 & $\times$ 8 & $\times$ 32 &    54.6  \\
        & 8 & $\times$ 16 & $\times$ 64 &   51.3   \\

    \bottomrule
    \end{tabular}
    \caption{Semantic segmentation on S3DIS.}
    \label{tab:quantized_S3DIS_results}
\end{table}

%-------------------------------------------------------------------------------------------------------------
\subsection{Ablation study}

We delve on the influence of the number of wavelet levels on the accuracy for the ModelNet10 \cite{wu20153d} dataset. All configurations use 8 bit quantization, and $\times 8$ wavelet compression, and we compare one, three and five levels in Eq. \eqref{eq:waveletHierarchy}. Table \ref{tab:ablationLevels} summarizes the reuslts and shows that three levels are the best option for this scenario.

\begin{table}
    \centering
    \begin{tabular}{ccccc}
    Model & Act.  & Wav.   & Wav.  & Overall   \\
     &  bits &  comp.  &  levels & acc. $\%$   \\
        \toprule
        WDGCNN\textsubscript{V \textsubscript{1}}
        & 8 & $\times$  8 & $\times$ 1 & 93.1     \\
        & 8 & $\times$ 8 & $\times$ 3 & 93.7     \\
        & 8 & $\times$ 8 & $\times$ 5 & 92.6    \\
\bottomrule
    \end{tabular}
    \caption{The impact of number of wavelet levels on ModelNet10 shape classification.}
    \label{tab:ablationLevels}
\end{table}

\iffalse
Modelnet V1, V2, V3
\begin{table}[]
    \centering
    \begin{tabular}{ccccc}
    model & 32W32A & 8W8A & 4W4A \\
    \toprule
       DGCNN V1  & - & - & - \\
       DGCNN V2  & - & - & - \\
       DGCNN V3  & - & - & - \\
   \midrule
       DGCNN V1 + wavelet  &  & - & - & - \\
       DGCNN V2 + wavelet  &  & - & - & - \\
       DGCNN V3 + wavelet  &  & - & - & - \\
    \bottomrule
    \end{tabular}
    \caption{Ablation study}
    \label{tab:ablation}
\end{table}

\fi

%-------------------------------------------------------------------------------------------------------------
\section{Conclusion}
We propose an efficient compression method for GCNs, based on the combination of compressed Haar wavelet and quantization methods. We conduct an extensive set of experiments for both geometrical and non-geometrical graph datasets, where it is demonstrated that our method is often associated with large performance gains, both in accuracy and computational efficiency terms. Furthermore, we demonstrate that 8-bit and 4-bit quantization can be applied to further reduce memory and computational costs, with minimal loss in performance, also while using baseline methods.

Our method has the potential to reduce inference times, memory  and computational costs related to deploying GCNs in real world applications, such as LiDAR-based point cloud segmentation for autonomous vehicles. Additionally, our method enables reduced training costs, as compressed wavelet-based convolution decreases the dimensions of the input tensor.

In our future work we plan to see if the low-precision performance can be improved using more advanced quantization schemes such as non-uniform and dynamic methods (which use different bit allocations for different layers), as well as topology aware quantization. Alternatively, this can be combined with the various weight pruning methods for further savings, as our compressed feature maps are dense.

% or


%\appendices
%\section{}

\section*{Acknowledgment}
The research reported in this paper was supported by the Israel Innovation Authority through Avatar
consortium.%, and by grant no. 2018209 from the United States - Israel Binational Science Foundation (BSF), Jerusalem, Israel. 
 ME is supported by Kreitman High-tech scholarship.

The authors thank Mr. Tal Kopetz for valuable conversations on the efficiency of CNNs on dedicated edge hardware.   

\ifCLASSOPTIONcaptionsoff
  \newpage
\fi

\bibliographystyle{IEEEtran}  

\bibliography{references}


%\end{IEEEbiography}



\end{document}